\documentclass[letterpaper, 10 pt, conference]{ieeeconf}  % Comment this line out if you need a4paper
\usepackage[utf8]{inputenc}
\usepackage{amsmath, amsfonts}
\usepackage{graphicx}
\usepackage{xcolor}
\usepackage{wrapfig}
\usepackage{adjustbox}
\usepackage{xspace}
\usepackage{epstopdf}
\usepackage{xcolor}
\usepackage{float}
\usepackage{soul}
\usepackage{hyperref}
\usepackage{siunitx}
\usepackage{booktabs}
\usepackage{algorithm}
\usepackage{algcompatible}
\usepackage[capitalize]{cleveref}
\usepackage{booktabs}
\usepackage{multirow}
\usepackage[para]{threeparttable} 
\usepackage{array}
\usepackage{colortbl}
\usepackage{siunitx}
\usepackage{pifont}

\usepackage{eqexpl}
\eqexplSetDelim{=}

\usepackage{algpseudocode}
\usepackage{array}
\usepackage{tabularx}
\usepackage[nolist]{acronym} % acronyms by the \ac{label} command

\usepackage{wrapfig}
\usepackage[noadjust]{cite}
\usepackage{comment}
\usepackage{censor}

\IEEEoverridecommandlockouts % <- needed for \thanks command

\newcommand{\systemname}{\textit{}\xspace}

\newacro{FDM}[FDM]{Fused Deposition Modelling}
\newacro{EIT}[EIT]{Electrical impedance tomography}
\newacro{SLA}[SLA]{Stereolithography}

\StopCensoring

 \usepackage[firstpageonly=true]{draftwatermark}

\SetWatermarkAngle{0}
\SetWatermarkColor{black}
\SetWatermarkLightness{0.5}
\SetWatermarkFontSize{9pt}
\SetWatermarkVerCenter{30pt}
\SetWatermarkText{\parbox{30cm}{%
\centering This work has been submitted to the IEEE for possible publication. \\
\centering Copyright may be transferred without notice, after which this version may no longer be accessible.
}}

\title{\systemname Toward Geometry-Scalable Whole-Body Touch for Humanoids: A 3D-Printed Conformal EIT Skin}

\xblackout{
\author{Haofeng Chen$^{1}$, Carson Kohlbrenner$^{2}$, Jiri Kubik$^{1}$, Lukas Rustler$^{1}$, Alexander Dickhans$^{2}$, \\ Karel Bartunek$^{1}$  Alessandro Roncone$^{2}$, Hyosang Lee$^{3}$, and Matej Hoffmann$^{1}$
\thanks{The authors are with $^{1}$Department of Cybernetics, Faculty of Electrical Engineering, Czech Technical University in Prague \{{\tt\small matej.hoffmann@fel.cvut.cz}\}, $^{2}$ Human Interaction and Robotics group at CU Boulder \{{\tt\small first.last@colorado.edu}\}, and $^{3}$Eindhoven University of Technology. H.C., L.R., and M.H. were supported by the European Union under the project Robotics and Advanced Industrial Production (reg. no. CZ.02.01.01/00/22\_008/0004590). C.K., A.D., and A.R. were supported by the NSF FW-HTF-R grant \#2222952. We thank Benn Proper for assistance with 3D printing the robot face.}
}
}
\begin{document}
\maketitle

\begin{abstract}
Whole-body tactile sensing is a prerequisite for humanoids that operate in contact-rich human environments, but conventional taxel arrays scale poorly with surface area, wiring complexity, and robot-specific curvature. We present a conformal electrical impedance tomography tactile skin fabricated through a geometry-adaptable additive-manufacturing workflow. A flexible conductive TPU layer forms a continuous sensing domain, while contact-induced coupling with conductive patches produces boundary voltage changes that are reconstructed using a one-step Gauss–Newton EIT solver. We first characterize the electromechanical design space of the layered structure and show that low-resistance contact-enhancement patches and a porous conductive TPU sensing layer improve sensitivity while preserving printability. We then validate contact localization on a planar prototype, a curved U-shaped prototype, and a qualitative iCub-face-shaped geometry. The curved sensor achieves a mean localization error of 6 mm over 18 contact positions without supervised post-processing. These results suggest that additively manufactured tomographic skins can reduce the morphology-specific redesign burden for humanoid tactile coverage and provide a practical route toward large-area contact sensing for human-centered deployment.
\end{abstract}

\section{Introduction}

Humanoids are intended to operate in spaces designed for people, from industrial workspaces to care settings and homes. In these settings, physical contact is a source of information for whole-body control, manipulation, and social physical interaction \cite{chen2025dexforce}. 
Large-area tactile skins have therefore become a central enabling technology for contact-rich humanoids. However, scaling them across curved robot morphologies remains challenging, as existing systems commonly rely on dense or modular arrays of discrete sensing cells, requiring extensive interconnection, packaging, and geometry-specific integration \cite{zhao2022large,cheng2019comprehensive,zhou2023tacsuit}.

\ac{EIT} offers a scalable approach to contact sensing by reconstructing contact-induced conductivity changes in a continuous sensing domain from a limited number of electrodes \cite{park2024graph,park2022biomimetic}. \ac{EIT} skins have been demonstrated over large areas \cite{yang2025body,zheng2025large,chen2022large}. However, applying them to non-planar, curved, and robot-specific surfaces remains challenging due to the sensing layer, electrode arrangement, mechanical support, and forward model often requiring extensive hand tailoring for each new surface.

Additive manufacturing could reduce this morphology-specific fabrication burden by using the CAD geometry of a target robot surface to design conformal sensing and structural layers for direct fabrication. Existing 3D-printed tactile sensors, however, are predominantly based on discrete capacitive or piezoresistive sensing units \cite{moeinnia2024wireless,gong2021metasense,imranuddin2025enhancing,massaroni2024fully,kohlbrenner2026gentactprox}. Recent studies have incorporated 3D-printed components into EIT sensors. Huang et al. \cite{huang2025design} investigated patterned 3D-printed EIT skins on planar fingertip-scale samples. Park et al. \cite{park2024graph} demonstrated EIT sensing on a humanoid-face geometry using a printed support followed by spray coating, molding, manually positioned conductive patches, and electrode assembly. Chen et al. \cite{CHEN2026118714} developed a printed conductive-PLA EIT sensor for a customized curved interface, focusing on anisotropic conductivity correction and interaction-pressure monitoring. These studies establish the feasibility of printed and non-planar EIT sensing, but the design of a flexible, printing-compatible continuous-domain skin that can be transferred across geometries remains under explored. A preliminary version of the curved sensor design and localization experiment was presented at the ICRA 2026 Workshop on Towards Large-Area Tactile Sensing Skins \cite{chen2026workshop}. The present paper substantially extends that work through electromechanical design characterization, planar localization experiments, force-response and multi-contact evaluations, and a proof-of-concept demonstration on an iCub-face geometry.

In this work, we present a conformal EIT tactile skin for contact localization on planar and curved surfaces. The flexible conductive TPU sensing layer forms an electrically continuous sensing surface, allowing contacts at arbitrary locations within the covered area to be localized without embedding and wiring a dense array of predefined taxels. EIT reconstructs the spatial conductivity perturbation from pairwise voltage changes acquired using a limited number of distributed electrodes. The sensor further combines 3D-printed structural layers and conductive contact-enhancement patches to enable conformal implementation on both flat and curved geometries.

The main contributions of this work are:
\begin{itemize}
    \item A conformal 3D-printed EIT tactile-skin architecture that combines a flexible conductive TPU sensing layer, printed structural layers, electrodes, and low-resistance contact-enhancement patches to form a continuous sensing domain on planar and curved surfaces.
    \item Experimental characterization of contact-patch material, conductive-layer thickness, and porosity, for geometry-aware contact reconstruction.
    \item Quantitative contact-localization evaluation on planar and U-shaped sensors, supplemented by multi-contact, force-response, and humanoid-face demonstrations.
\end{itemize}
Together, these results provide a basis for extending additively manufactured EIT tactile skins to larger and more complex robot surfaces, with future work focusing on automated geometry-to-sensor design, fully printed electrical interfaces, and whole-body integration.

\section{Methods}
\label{Methods}
\subsection{Sensor Design and Fabrication}
\label{sec:material_selection}
\begin{figure}
    \centering
    \includegraphics[width=0.99\linewidth]{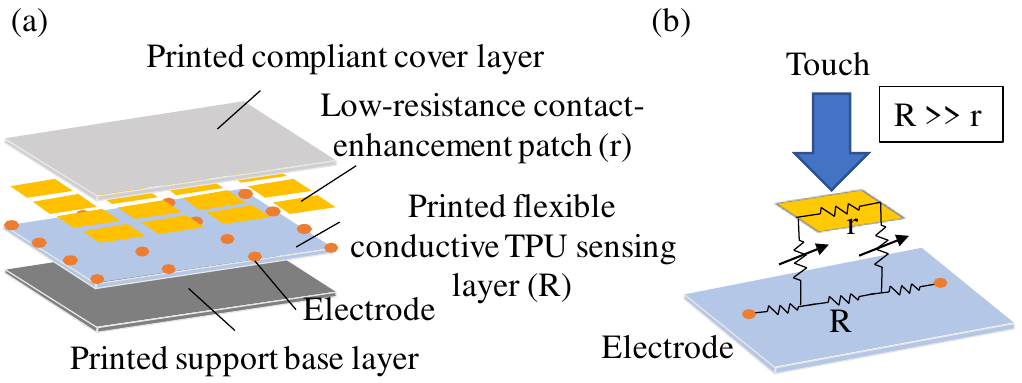}
    \caption{Schematic of the proposed conformal EIT tactile skin. (a) Layered sensor structure. (b) Touch increases local coupling between the low-resistance contact-enhancement patches (r) and the higher-resistance flexible conductive TPU sensing layer (R, with R $\gg$ r).}
    \label{fig:sensor}
\end{figure}
The sensor is designed as a conformal EIT tactile skin for curved robotic surfaces. As illustrated in Fig.~\ref{fig:sensor} (a), the sensor follows a layered structure consisting of a flexible conductive TPU sensing layer, contact-enhancement patches, a printed support base layer, and a printed compliant cover layer. The sensing mechanism is illustrated in Fig.~\ref{fig:sensor}(b).  
When a normal load is applied, the effective contact area between the contact-enhancement patch and the conductive sensing layer increases, reducing the local interfacial resistance and producing a conductivity perturbation in the sensing domain. 
This perturbation is measured as boundary voltage changes and reconstructed using EIT.

\begin{figure} %H为当前位置，!htb为忽略美学标准，htbp为浮动图形
\centering %图片居中
\includegraphics[width=0.48\textwidth]{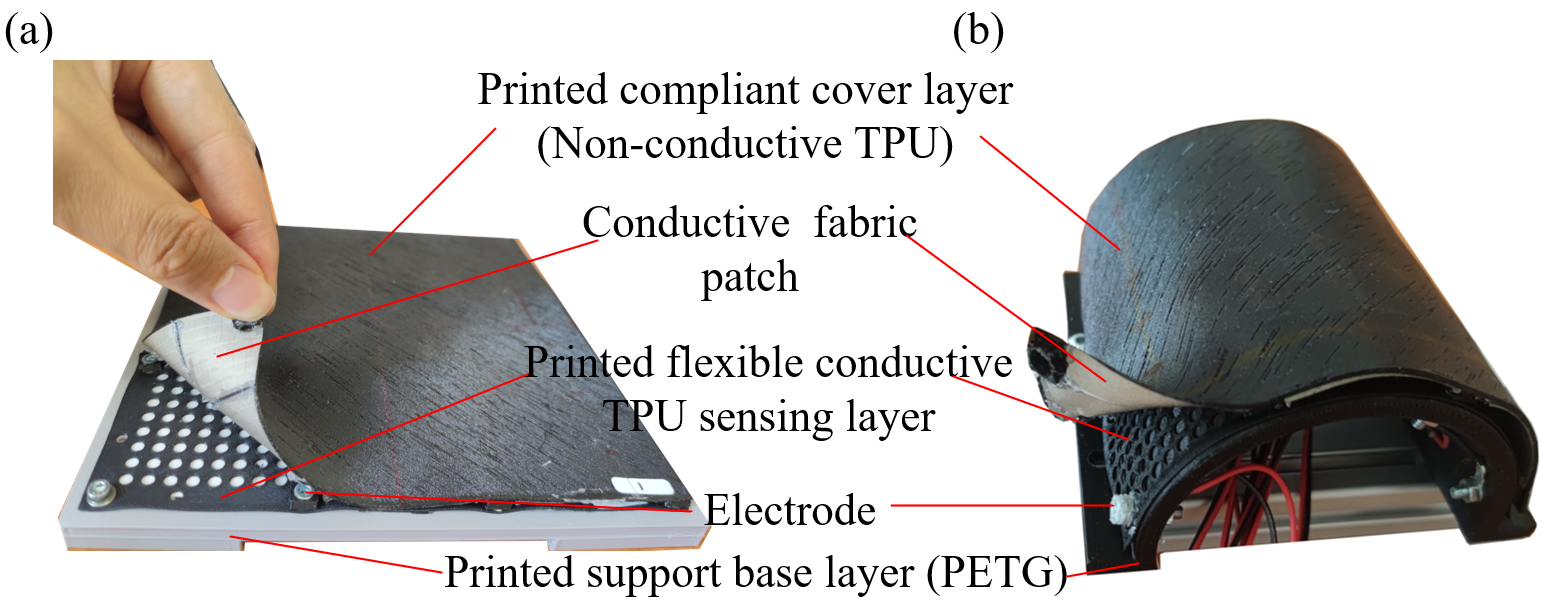} %插入图片，[]中设置图片大小，{}中是图片文件名
%\caption{Planar sensor shape and curved U-shaped sensor.} %
\caption{Fabricated sensor samples. The planar sensor is on the left, the U-shaped sensor is on the right. Both sensors follow the same layer concept, with the conductive fabric patches adhered to the cover layer, while the patch geometry was adapted to each sensor shape.} %
\label{fig:tactile_sensor} %用于文内引用的标签
\end{figure}

Fig.~\ref{fig:tactile_sensor} shows the fabricated planar and curved EIT tactile sensor prototypes. The planar sensing area is 150 mm $\times$ 150 mm, while the U-shaped sensor has a radius of 40 mm and a length of 100~mm.
Both prototypes use Recreus Conductive Filaflex TPU (Recreus, Spain) for the conductive sensing layer, with a sensing-layer thickness of 0.4~mm. Filaflex 60A TPU (Recreus, Spain) was used for the compliant cover because of its flexibility and compatibility with \ac{FDM} printing.
Conductive fabric patches cut from WE-TS Shielding Textiles 33025 (W\"urth Elektronik eiSos, Germany) were used as contact-enhancement patches. 
The porosity values of the conductive TPU sensing layers were 26.96\% and 38.40\% for the planar and U-shaped sensors, respectively.
The complete stack was mounted on a 3D-printed PETG base to define the sensor geometry and support both planar and curved prototypes. Both prototypes used sixteen electrodes arranged in a $4\times4$ distributed layout over the sensing surface. The distributed layout was selected to provide measurement coverage across the full tactile surface, including the interior region, following prior EIT/ERT tactile-sensing studies that used internal electrodes to improve central-region sensitivity and spatial resolution\cite{Lee2019Internal,park2024graph}.

Design parameters such as contact-enhancement patch material, conductive TPU sensing layer thickness, and conductive TPU sensing layer porosity were selected according to the electromechanical characterization described in Section \ref{Electromechanical} and layer-specific print settings are reported in Table~\ref{tab:printing_parameters}.

The planar sensor is used for controlled quantitative evaluation, including single-contact localization, force-dependent response, and multi-contact reconstruction. The U-shaped sensor is used to validate the transferability of the sensor fabrication and contact reconstruction pipelines to a curved surface.

\begin{table}[t]
\centering
\caption{Printing parameters for each filament.}
\label{tab:printing_parameters}
\begin{tabular}{lccc}
\toprule
\textbf{Filament} & \textbf{Nozzle } & \textbf{Bed} & \textbf{Speed} \\
 & \textbf{ ($^\circ$C)} & \textbf{ ($^\circ$C)} & \textbf{(mm/s)} \\
\midrule
Conductive Filaflex & 250 & 50 & 15 \\
Filaflex 60A TPU    & 235 & 30 & 15 \\
PETG                & 230 & 70 & 30 \\
\bottomrule
\end{tabular}
\vspace{-2em}
\end{table}

\subsection{Inverse Reconstruction}

The relationship between boundary voltages $\mathbf{V} \in \mathbb{R}^m$ and conductivity distribution $\boldsymbol{\sigma} \in \mathbb{R}^n$ is governed by the forward operator $F(\cdot)$, i.e., $\mathbf{V} = F(\boldsymbol{\sigma})$. Since $F(\cdot)$ is inherently nonlinear, a linearized differential formulation is commonly adopted, relating the conductivity change $\Delta\boldsymbol{\sigma} \in \mathbb{R}^n$ to the boundary voltage change $\Delta\mathbf{V} \in \mathbb{R}^m$ as
\begin{equation}
    \Delta\mathbf{V} = \mathbf{J}\,\Delta\boldsymbol{\sigma} + \mathbf{n}_{noise}
\end{equation}
where $\mathbf{J} \in \mathbb{R}^{m \times n}$ is the sensitivity matrix, $\mathbf{n}_{noise}$ denotes the noise matrix \cite{Adler2007Temporal}.

Following the one-step Gauss--Newton formulation~\cite{adler1996electrical}, the conductivity change is recovered in closed form as
\begin{equation}
    \Delta\boldsymbol{\sigma} = \underbrace{(\mathbf{J}^T\mathbf{W}\mathbf{J} + \lambda^2\mathbf{R})^{-1}\mathbf{J}^T\mathbf{W}}_{\mathbf{RM}}\,\Delta\mathbf{V}
\end{equation}
where $\mathbf{J} \in \mathbb{R}^{m \times n}$ is the sensitivity matrix relating boundary voltage changes $\Delta\mathbf{V} \in \mathbb{R}^m$ to conductivity changes $\Delta\boldsymbol{\sigma} \in \mathbb{R}^n$, $\mathbf{W}$ is a weighting matrix, $\lambda$ is a regularization hyperparameter, and $\mathbf{R}$ is computed using the Laplace prior~\cite{RN644}. The reconstruction matrix $\mathbf{RM}$ is precomputed offline, enabling real-time tactile reconstruction at deployment.

\subsection{Contact Localization}
To extract the estimated contact location, the reconstructed image was thresholded to suppress background noise, retaining only elements satisfying
\begin{equation}
    \Omega_c = { i \mid \Delta\sigma_i > 0,\ \Delta\sigma_i > \alpha \max(\Delta\sigma) },
    \label{thresholded_sigma}
\end{equation}
where $\Delta\sigma$  is the reconstructed conductivity change of element $i$ and $\alpha$ is the threshold ratio. The estimated contact position was then calculated as the weighted centroid of the retained region:
\begin{equation}
\hat{x}=\frac{\sum_{i \in \Omega_c} x_i \Delta\sigma_i}{\sum_{i \in \Omega_c} \Delta\sigma_i},
\qquad
\hat{y}=\frac{\sum_{i \in \Omega_c} y_i \Delta\sigma_i}{\sum_{i \in \Omega_c} \Delta\sigma_i},
\end{equation}
where $(x_i,y_i)$ denotes the center coordinate of element $i$. The localization error was defined as the Euclidean distance between the true contact position $(x,y)$ and the estimated position $(\hat{x},\hat{y})$:

\begin{equation}
    e = \sqrt{(\hat{x}-x)^2+(\hat{y}-y)^2}.
\end{equation}

For the curved surface sensor, the raw contact estimate was then computed as the conductivity- and volume-weighted centroid of the active elements:
\begin{equation}
\hat{\mathbf{p}} =
\frac{
\sum_{i \in \mathcal{S}} \delta\sigma_i V_i \mathbf{c}i
}{
\sum{i \in \mathcal{S}} \delta\sigma_i V_i
},
\label{eq}
\end{equation}
\begin{figure} 
\centering
\includegraphics[width=0.45\textwidth]{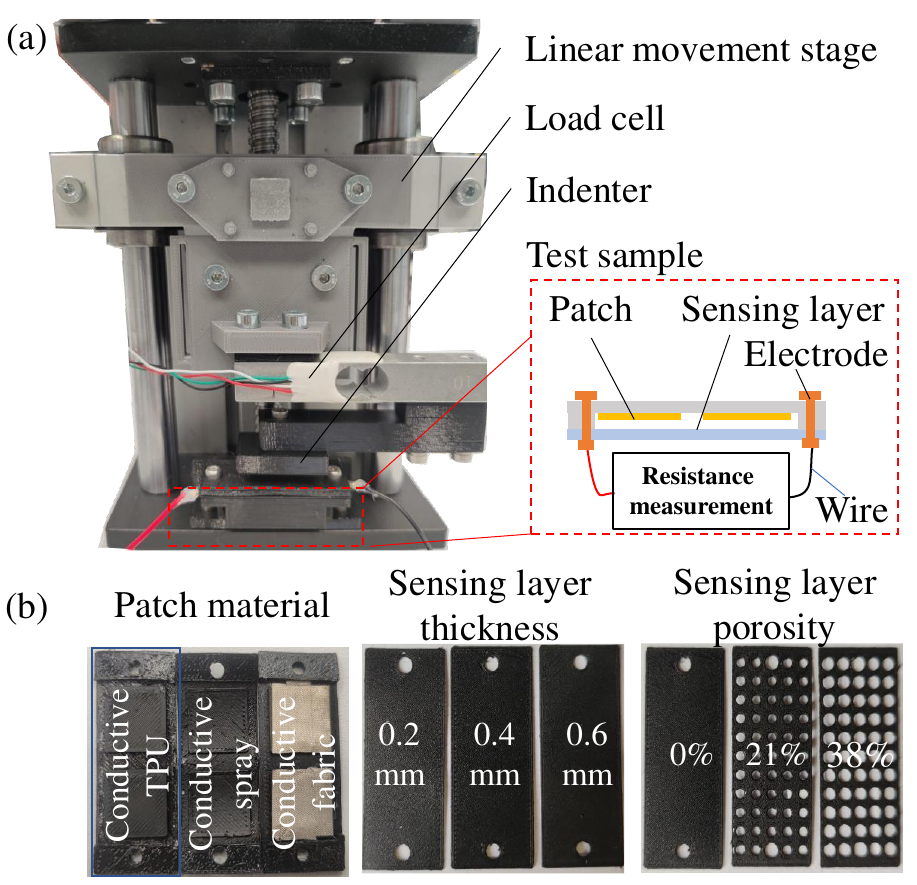} 
\caption{Experimental setup and samples for electromechanical characterization. (a) Normal-loading setup and resistance measurement path.  (b) Sample groups used to compare contact-enhancement patch material, conductive TPU sensing layer thickness, and conductive TPU sensing layer porosity. } %
\label{linear_setup}
\end{figure}

where $\delta\sigma_i$, $V_i$, and $\mathbf{c}_i$ denote the reconstructed conductivity change, volume, and centroid of the $i$-th tetrahedral element, respectively.

Because the weighted centroid may lie inside the volumetric finite-element mesh rather than exactly on the physical tactile surface, the estimated position was projected onto the nearest surface mesh node. Surface nodes were obtained by extracting boundary faces that belong to only one tetrahedral element. The projected estimate is denoted as $\hat{\mathbf{p}}^{\mathrm{surf}}$. The ground-truth indentation point $\mathbf{p}^{\mathrm{gt}}$was defined from the manually measured contact location on the sensor surface.
\begin{equation}
\mathrm{PE}
=
\left\|
\hat{\mathbf{p}}^{\mathrm{surf}}
-
\mathbf{p}^{\mathrm{gt}}
\right\|_2 .
\label{eq:pe}
\end{equation}
This projection-based Euclidean error was used instead of a graph-based surface geodesic distance because the finite-element model represents the sensor as a finite-thickness volumetric shell. The extracted boundary mesh therefore contains multiple boundary layers and triangulation-dependent edge connections. A shortest-path geodesic computed on this mesh can be affected by mesh resolution, triangle orientation, and projection ambiguity between neighboring surface layers. In contrast, the projected Euclidean distance directly measures the spatial discrepancy between the estimated and true contact locations on the physical sensor surface, while remaining simple, reproducible, and independent of mesh-edge connectivity.

\section{Results}
\subsection{Electromechanical Characterization}
\label{Electromechanical}

\begin{table}[t]
\centering
\footnotesize
\caption{Sample configurations for comparing design factors in electromechanical characterization.}
\label{tab:electromechanical_samples}
\renewcommand{\arraystretch}{1.18}
\setlength{\tabcolsep}{3pt}
\begin{tabularx}{\columnwidth}{p{0.25\columnwidth} X X}
\toprule
\textbf{Varied factor} & \textbf{Contact-enhancement patch} & \textbf{Conductive TPU sensing layer} \\
\midrule

Patch material 
& cTPU (1.5 mm) / conductive spray / conductive fabric 
& cTPU, 0.4 mm,  0\% porosity \\

Sensing layer thickness
& Conductive fabric patch 
& cTPU: 0.2, 0.4, 0.6 mm, 0\% porosity \\

Sensing layer porosity 
& Conductive fabric patch 
& cTPU, 0.4 mm: 0, 21, 38\% porosity \\
\bottomrule
\vspace{1mm}
\end{tabularx}
\footnotesize \parbox{\columnwidth}{ \textit{Note:} cTPU denotes 3D-printed Recreus Conductive Filaflex TPU (Recreus, Spain). 
}
\vspace{-1mm}
\end{table}

\begin{figure*}[htbp]
\centering
\includegraphics[width=0.98\textwidth]{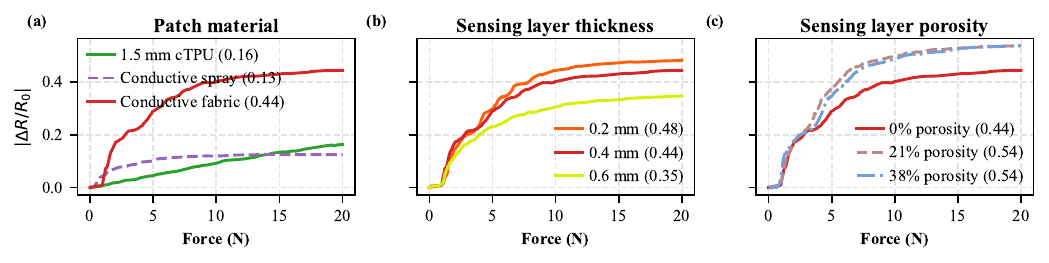}
\caption{Comparison of electromechanical responses for fabrication-relevant design factors under normal loading from 0 to 20 N. (a)  patch material, (b) conductive TPU sensing layer thickness, and (c) conductive TPU sensing layer porosity. The curves show the force-dependent normalized resistance change. 
The values in parentheses indicate the maximum $|\Delta R/R\_0|$ observed over the 0--20\ N loading range.} %
\label{Sensitivity_Comparison}
\end{figure*}

Before fabricating the EIT tactile-skin prototypes, planar samples were used to identify the dominant design factors affecting the electromechanical response of the layered sensing structure. The goal of this characterization was not to exhaustively optimize all geometric parameters, but to derive practical design principles that balance sensitivity, printability, and mechanical robustness.

The experimental setup is shown in Fig.~\ref{linear_setup} (a). Each sample was compressed using a linear movement stage equipped with a load cell and a rectangular indenter 
(30 mm $\times$ 10 mm $\times$ 6 mm). The sample resistance was measured between two electrodes connected to the conductive TPU sensing layer, as illustrated in the inset of Fig.~\ref{linear_setup}(a). Each sample measured $50 \times 20$~mm, with an electrode spacing of 40~mm.
For each test, the no-load resistance was defined as $R_0 = R(0)$, and the normalized resistance change at each force level was computed as $|\Delta R/R_0| = \frac{|R(F)-R_0|}{R_0}$, where $R(F)$ is the resistance under the applied normal force $F$.
Representative sample groups are shown in Fig.~\ref{linear_setup} (b), and the corresponding test configurations are summarized in Table~\ref{tab:electromechanical_samples}. For the patch-material comparison, all contact patches had the same footprint, consisting of two 15~mm $\times$ 15~mm square patches. The cTPU patch was directly 3D printed with a thickness of 1.5~mm. 
For the conductive spray and conductive fabric, 1.5~mm non-conductive TPU patches with the same footprint were first printed as mechanical substrates, and the conductive spray or conductive fabric was then applied on top. Three coats of conductive spray paint were applied to the sample before testing. The non-conductive TPU patch used Filaflex 60A TPU (Recreus, Spain).  The conductive spray used 838AR conductive spray paint (MG Chemicals, Canada), while the conductive fabric was cut from WE-TS Shielding Textiles 33025 (W\"urth Elektronik eiSos, Germany). 

The characterization followed the design logic illustrated in Table.~\ref{tab:electromechanical_samples}. The comparison of electromechanical responses for fabrication-relevant design factors is shown in Fig. \ref{Sensitivity_Comparison}.

\subsubsection{Patch materials} 
The selected patch materials have different manufacturer-reported electrical properties: the conductive spray paint has a nominal sheet resistance of 50~$\Omega$/sq, the conductive fabric has a reported surface resistance of 0.04~$\Omega$/cm$^2$, and the printed cTPU has an electrical resistivity of 0.9--16.5 $\Omega \cdot$ cm. 
Since these values are reported using different measurement methods and units, they are treated only as indicative material properties rather than directly comparable quantities.

Fig.~\ref{Sensitivity_Comparison}(a) compares the effect of contact-enhancement patch material. 
The conductive fabric patch produced the largest normalized resistance change, while the printed cTPU patch showed the weakest response. The conductive spray patch exhibits an earlier low-force response than the printed cTPU patch, but its maximum resistance change remains limited. This may be caused by non-uniform coating on the rough surface of the 3D-printed TPU substrate, which can increase the effective interfacial resistance. These results indicate that a low-resistance contact-enhancement patch is important for producing a strong local coupling with the conductive TPU sensing layer. 
Therefore, conductive fabric patches were used in the following sensor prototypes.

\subsubsection{Sensing layer thickness} After selecting the contact-enhancement patch material, we evaluated the thickness of the conductive TPU sensing layer. As shown in Fig.~\ref{Sensitivity_Comparison} (b), thinner sensing layers produced larger normalized resistance changes. This is because reducing the sensing layer thickness increases the effective resistance of the conductive TPU layer, making the local contact-induced perturbation more visible in the measured resistance response. However, the 0.2 mm layer was more difficult to print and handle reliably. 
Therefore, a 0.4~mm conductive TPU sensing layer was selected as a compromise between sensitivity, printability, and mechanical robustness.

\subsubsection{Sensing layer porosity} Finally, we evaluated the porosity of the conductive TPU sensing layer. Three 0.4~mm sensing layers were fabricated: 0\% porosity, 21\% porosity, and 38\% porosity. Porosity was introduced into the samples via grid-aligned spreading of specifically sized holes during design.
As shown in Fig.~\ref{Sensitivity_Comparison}(c), introducing porosity increased the normalized resistance change compared with the solid layer (0\% porosity). 
This suggests that porosity can increase the effective resistance of the sensing layer while preserving a printable layer thickness. 
Further increasing the porosity from 21\% to 38\% did not noticeably increase the maximum response, indicating that moderate porosity is sufficient for improving sensitivity in this design.

Overall, these results indicate two practical design principles for the proposed tactile skin. 
First, the contact-enhancement patch should provide a low-resistance local interface relative to the conductive TPU sensing layer. 
Second, the conductive TPU sensing layer should have sufficiently high effective resistance to yield measurable resistance changes, but this must be balanced against printability and mechanical robustness. 
The final sensor design follows these principles while selecting parameters that remain feasible for conformal fabrication on planar and curved surfaces.

\subsection{Single-Contact Localization on the Planar Sensor}

\begin{figure} 
\centering
\includegraphics[width=0.48\textwidth]{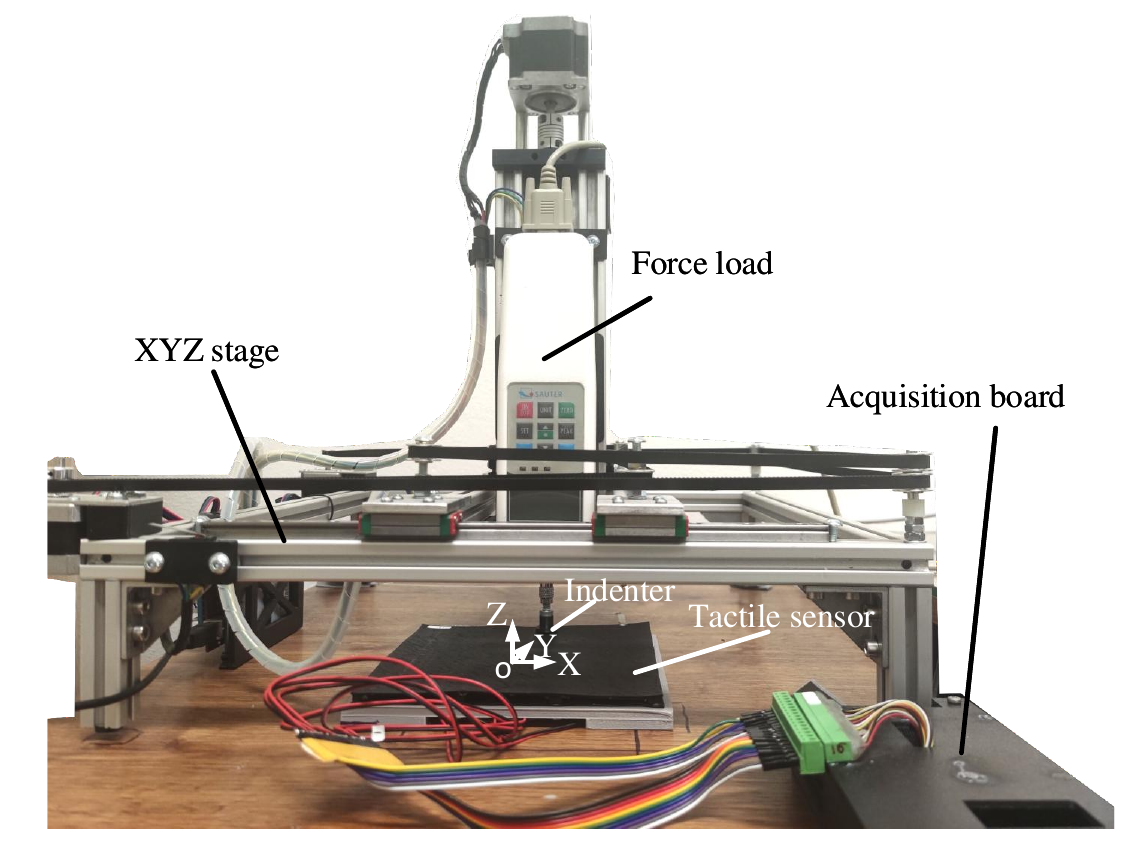}
\caption{Experimental setup for planar single-contact localization.} %
\label{fig:xyz_setup} 
\end{figure}

The planar sensor was first evaluated under controlled single-contact indentation. 
As shown in Fig.~\ref{fig:xyz_setup}, the sensor was fixed under a XYZ positioning stage, and a force gauge with a flat-tip cylindrical indenter was used to apply normal contact at predefined locations. The indenter diameter was 15~mm, providing a consistent contact area across tests.
This setup allowed the contact position and loading level to be controlled while electrode voltage measurements were recorded by the EIT acquisition board. A two-terminal measurement scheme was used, where current injection and voltage measurement were performed simultaneously for each electrode pair. With 16 electrodes, 120 independent electrode-pair measurements were acquired per frame. Details of the acquisition circuitry can be found in \cite{chen2022large}.

\begin{figure}
\centering
\includegraphics[width=0.25\textwidth]{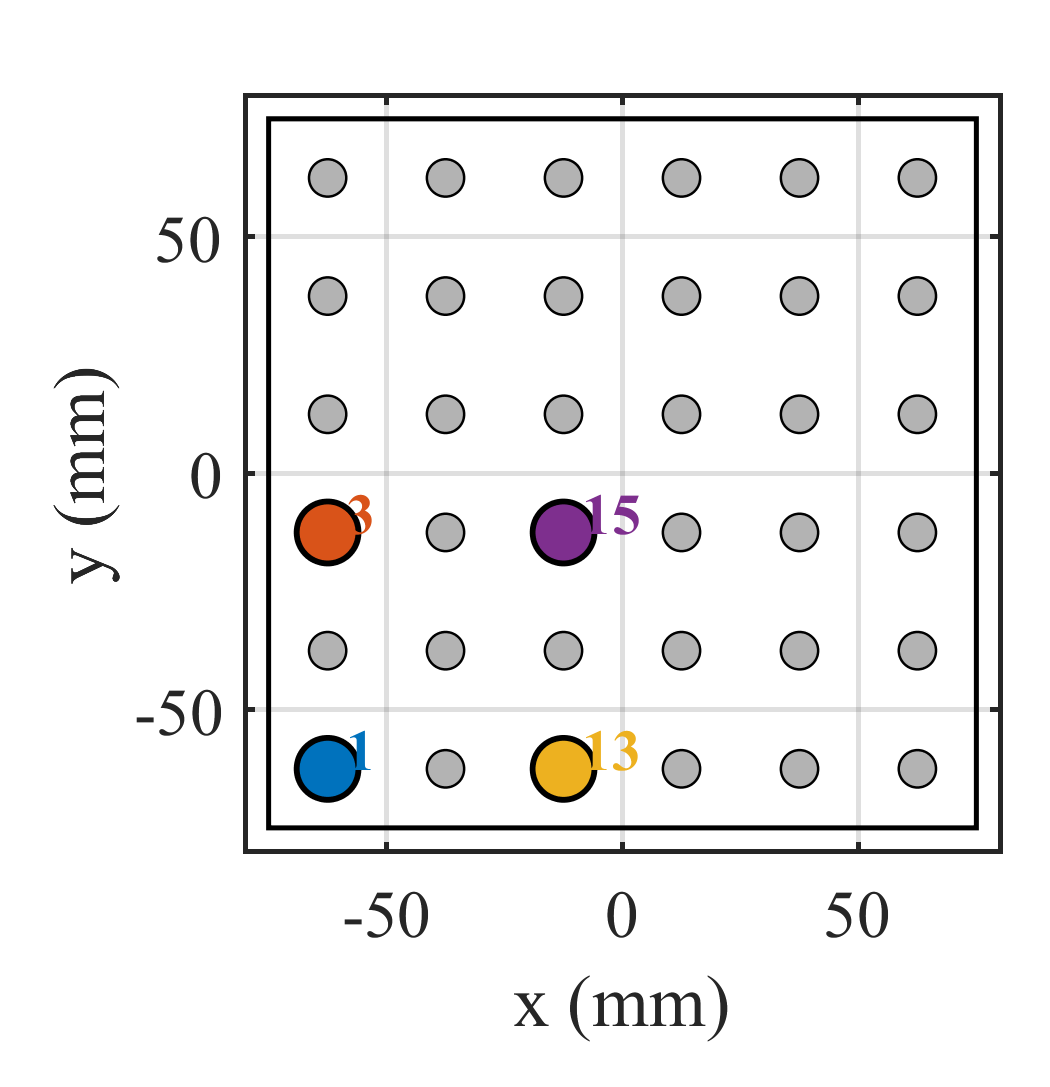}
\caption{Schematic diagram of single the touch test with predefined contact positions 1, 3, 13, and 15 highlighted.} %
\label{fig:representative_positions} 
\end{figure}
Fig.~\ref{fig:representative_positions} shows the predefined contact positions used for the single-touch test. 
The positions were distributed across the sensing area to evaluate localization performance at both central and near-boundary regions. 
For each indentation, the voltage change relative to the no-contact reference was reconstructed into a conductivity-change image using the one-step Gauss--Newton EIT solver.

\begin{figure} 
\centering 
\includegraphics[width=0.48\textwidth]{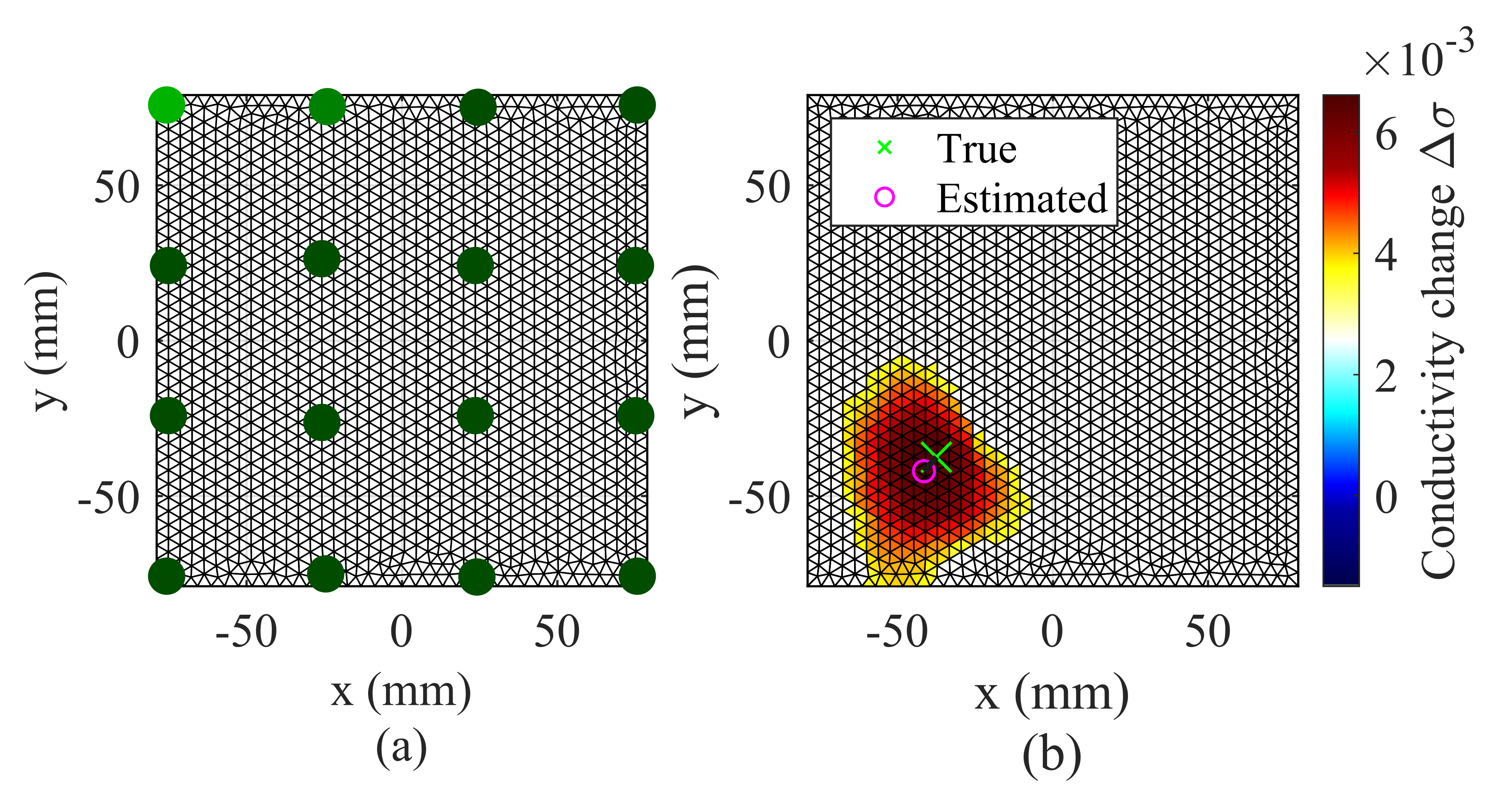} 
\caption{Illustration of the reconstruction and localization process.
(a) FEM model and electrode configuration of the EIT tactile sensor.
(b) Representative reconstructed conductivity-change image for a single contact. The true contact location and the extracted estimated position are marked, from which the XY localization error is calculated.} %
\label{fig:reconstruction_process} 
\end{figure}

The sensor geometry was designed in CAD software and imported into EIDORS to generate the finite element model (FEM), which served as the basis for both electrical simulation and EIT image reconstruction. 

As shown in Fig.~\ref{fig:reconstruction_process}, sixteen electrodes (green dots) were arranged in a $4\times4$ grid over the sensing area. 
Fig. \ref{fig:reconstruction_process} (b) shows a representative reconstructed conductivity-change image under a single contact, with the true and estimated contact positions indicated by green and magenta markers, respectively.

\begin{figure} 
\centering
\includegraphics[width=0.48\textwidth]{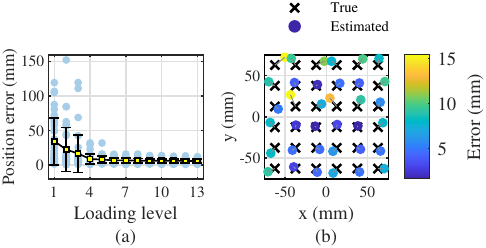} 
\caption{Localization performance of the 3D-printed EIT tactile sensor.(a) Position error under different loading levels. Markers and error bars indicate the mean and standard deviation. The loading level denotes the predefined indentation setting, not closed-loop force control; the corresponding load-cell measurements increased monotonically from approximately 1.5\~N to 16\~N across levels 1--13. (b) Representative localization map at loading level 6. Black crosses, colored circles, and gray lines denote true positions, estimated positions, and localization error vectors, respectively.} %
\label{fig:position_error} 
\end{figure}

Fig.~\ref{fig:position_error} summarizes the localization performance under different loading levels. The loading level represents the predefined experimental force setting, and the corresponding measured force increased monotonically from approximately 1.5 N to 16 N across levels 1--13. As shown in Fig.~\ref{fig:position_error}(a), the localization error decreases as the contact force increases. At low loading levels, the reconstructed response is weak and more strongly affected by measurement noise and the spatially non-uniform sensitivity of EIT, resulting in larger errors and higher variability. For loading level indices 1--3, the average localization error was $25 \pm 30$ mm. In contrast, for loading level indices 11--13, the error decreased to $6 \pm 3$ mm, indicating that the localization performance becomes stable once the contact-induced conductivity perturbation is sufficiently strong.

A representative localization result at force index 6 is shown in Fig.~\ref{fig:position_error} (b). At this force level, the mean localization error was $7 \pm 3.5$ mm over 36 valid contact samples, with a median error of 6.66 mm and a 90th percentile error of 10.83 mm. The estimated positions generally follow the spatial distribution of the true contact points, demonstrating that the proposed 3D-printed EIT sensor can recover contact locations over the planar sensing area.

\subsection{Force-Response Characterization}
\begin{figure} 
\centering
\includegraphics[width=0.45\textwidth]{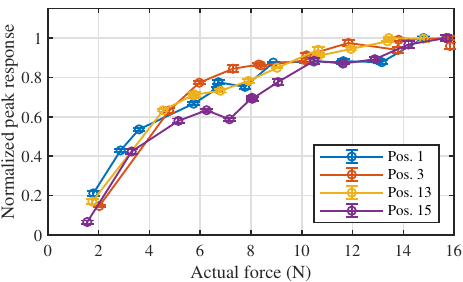} 
\caption{Reconstructed normalized peak response versus measured contact force at four representative positions. For each frame, the peak value was defined as the maximum positive value of the reconstructed conductivity change. 
At each position, the peak responses were normalized by the mean peak response measured at the highest force level, yielding a dimensionless normalized peak response.  Error bars indicate the standard deviation across valid frames within the selected pressing window. The normalized peak response increases with applied force and gradually approaches saturation at high force levels.} %
\label{fig:force_response}
\end{figure}

To examine how the reconstructed response varies with contact force, Fig. \ref{fig:force_response} shows the relationship between the measured contact force and the normalized peak response at four representative positions. For each frame, the peak response was defined as the maximum positive value in the reconstructed conductivity change $\Delta \sigma$. This scalar value represents the strongest local contact-induced conductivity perturbation in the conductive TPU sensing layer. 

Four representative positions were selected from different regions of the sensing area, as shown in Fig.~\ref{fig:representative_positions}, to examine whether the force-response trend was consistent across locations. The experiment was conducted using predefined loading levels, while the response curves in Fig.~\ref{fig:force_response} are plotted against the actual measured force. For each representative position, the peak response was first averaged over valid frames at each loading condition and then normalized by the mean peak response at the highest loading level of that position, yielding a dimensionless normalized peak response.  The normalized responses from the four positions were subsequently averaged to quantify the overall force-response trend. The forces corresponding to 50\%, 80\%, and 90\% of the maximum response were estimated by linear interpolation between adjacent measured force points.
Quantitatively, the average normalized peak value increased from $0.15 \pm 0.06$ at the lowest force level to $0.80 \pm 0.10 $ at an intermediate force level, and approached unity at the highest measured forces. The response reached 50\%, 80\%, and 90\% of its maximum at $3.8 \pm 0.4$ N, $8 \pm 1$ N, and $12 \pm 2$ N, respectively. Above 12 N, the normalized response remained close to saturation, with an average value of $0.97 \pm 0.04$.

This trend indicates that the EIT reconstruction contains information related to contact intensity. However, the response is not perfectly linear, which can be attributed to the nonlinear contact mechanics between the conductive patch and the conductive TPU sensing layer, as well as the saturation of the effective contact area under larger loads. Therefore, the current result demonstrates force-sensitive reconstruction behavior, while accurate force estimation would require position-dependent calibration or learning-based compensation.

\subsection{Multi-Contact Reconstruction}
\label{MultiContactReconstruction}
\begin{figure} 
\centering 
\includegraphics[width=0.48\textwidth]{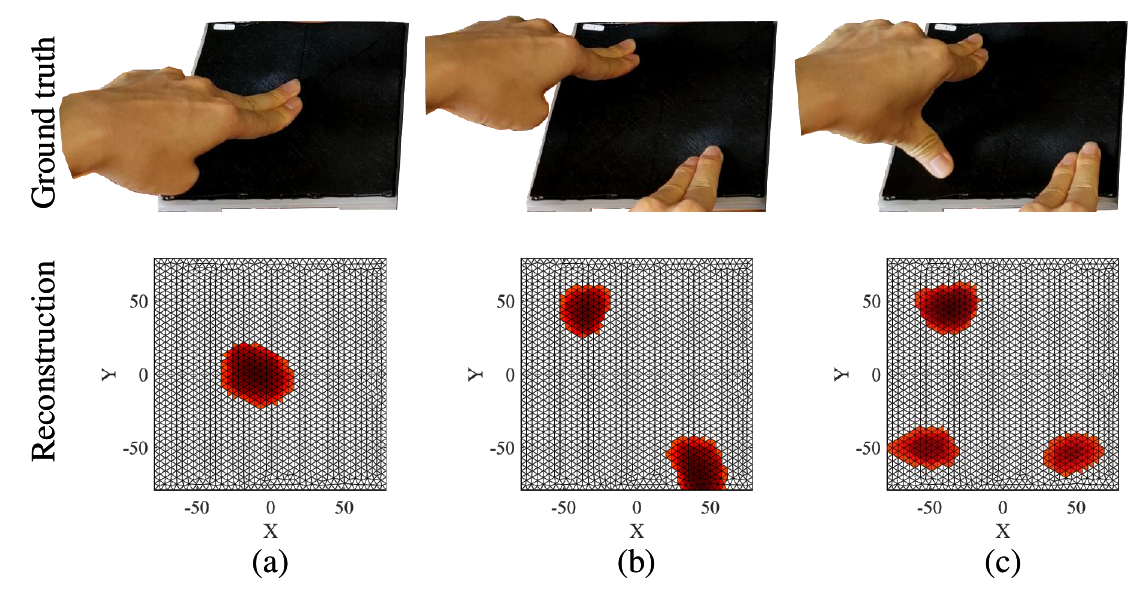} 
\caption{Multi-contact reconstruction on the planar sensor. 
The top row shows the applied contact conditions, and the bottom row shows the corresponding EIT reconstruction. 
The reconstructed hotspots are consistent with the applied contact regions for one-, two-, and three-contact cases.} %
\label{fig:square15cm_rec} 
\end{figure}

The planar sensor was further evaluated under multi-contact conditions to demonstrate the ability of EIT to capture spatially distributed touch. 
Fig.~\ref{fig:square15cm_rec} shows representative one-, two-, and three-contact cases. 
For each case, the top row shows the applied contact condition, while the bottom row shows the corresponding reconstructed conductivity-change image.
The reconstructed hotspots are spatially consistent with the applied contact regions, indicating that the sensor can distinguish multiple separated contacts on the same sensing surface. 
Compared with the single-contact case, the reconstructed regions become broader and partially blurred as the number of contacts increases. 
This is expected because EIT is a diffusive imaging modality with limited spatial resolution, and nearby conductivity perturbations may overlap in the reconstructed image. 
Nevertheless, the result demonstrates the potential of the proposed 3D-printed EIT tactile skin for distributed contact detection without requiring a dense taxel array.

\subsection{Localization on the Curved U-Shaped Sensor}
\begin{figure} 
\centering
\includegraphics[width=0.48\textwidth]{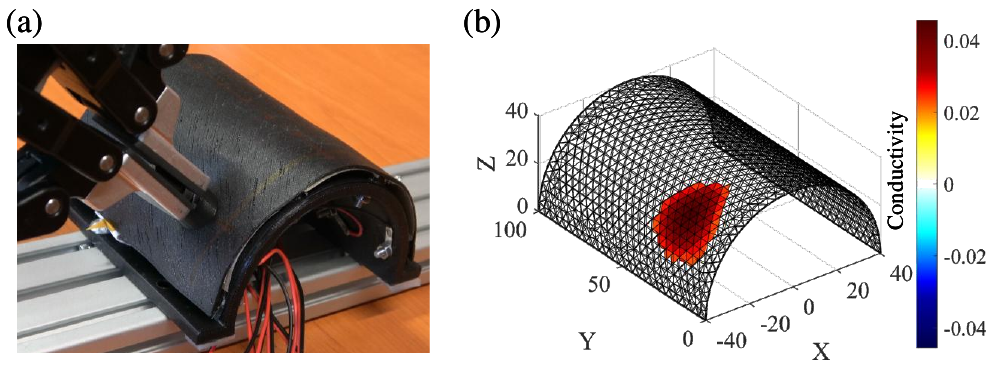} 
\caption{Experimental validation. (a) Photograph of the indentation test setup. (b) Representative EIT reconstruction showing the localized conductivity change under single-point contact.} %
\label{fig:ushape_reconstruction} 
\end{figure}

To evaluate whether the sensor design can be extended from a planar surface to a curved robotic surface, the U-shaped prototype was tested under controlled single-point indentation. The sensor adopts a multi-layer architecture as illustrated in Fig.~\ref{fig:tactile_sensor}. The U-shaped substrate has a radius of 40~mm and a length of 100~mm.

Fig.~\ref{fig:ushape_reconstruction} shows the experimental setup and a representative 3D EIT reconstruction. 
A 15~mm diameter indenter was mounted on the gripper of a Kinova Gen3 robotic arm and brought into contact with the curved tactile sensor under position control. 
The indentation was adjusted to obtain a stable reconstruction response, rather than to enforce the same contact force at all positions. 
Thus, this experiment evaluates curved-surface contact localization rather than force-controlled sensing performance. 

A total of 18 predefined indentation positions were tested on the curved surface. 
For each position, multiple frames around the peak response were averaged to reduce frame-to-frame noise before reconstruction. 
The estimated contact location was extracted from the reconstructed conductivity-change image using a thresholded weighted center-of-mass method.

\begin{figure}
\centering
\includegraphics[width=0.4\textwidth]{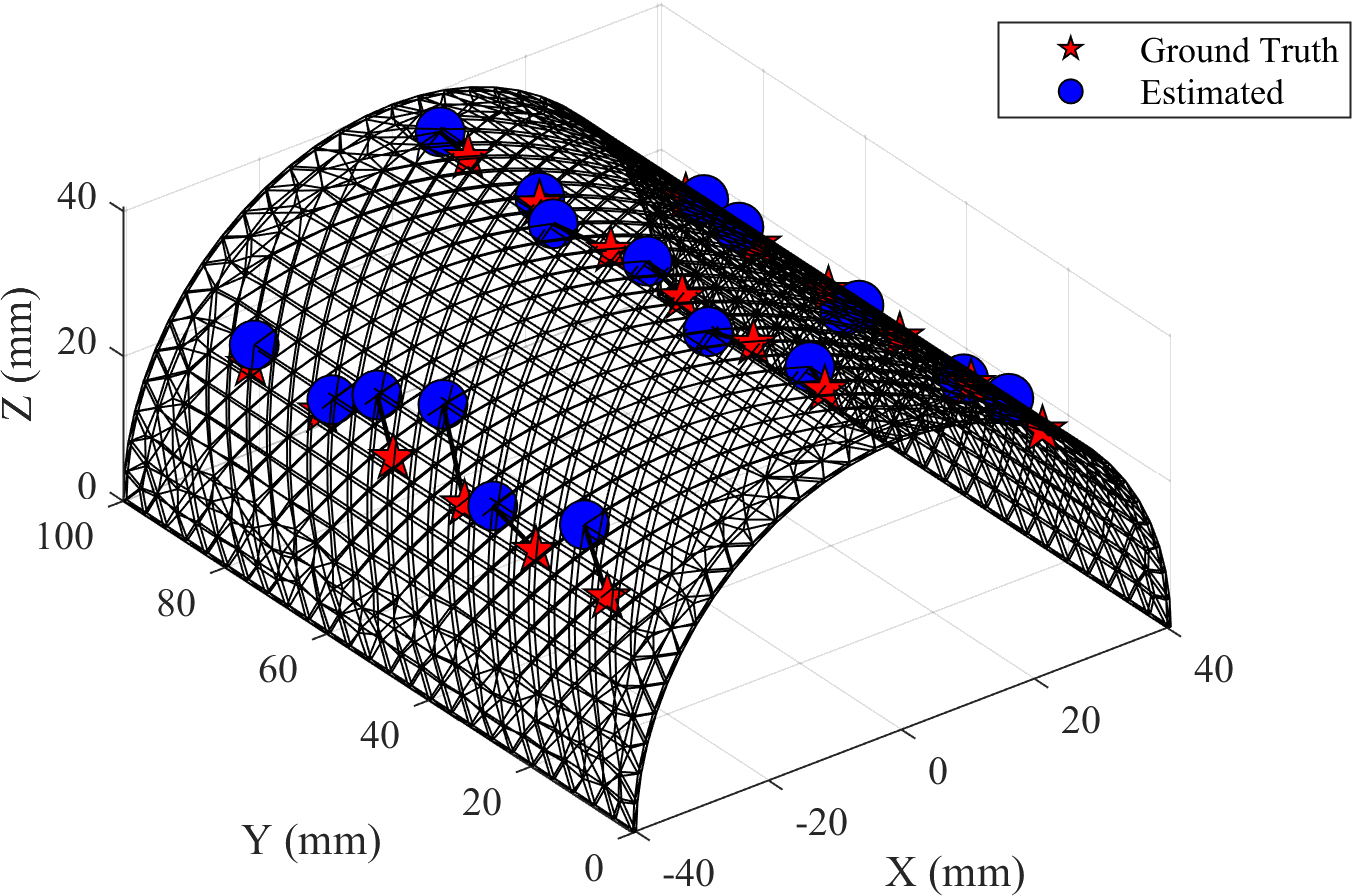} 
\caption{Spatial distribution of ground truth and estimated contact positions on the sensor surface.} %
\label{fig:ushape_position} 
\end{figure}

Fig.~\ref{fig:ushape_position} compares the ground-truth and estimated contact positions on the U-shaped sensor surface. 
The system achieved a mean localization error of \(6 \pm 4\)~mm across all 18 positions, with errors ranging from 1.5~mm to 14.4~mm. 
The result demonstrates that the proposed 3D-printed EIT sensor can localize contact on a curved surface with millimeter-level accuracy. 
Higher errors were mainly observed in regions with weaker boundary sensitivity and larger manual indentation uncertainty, suggesting that future work should improve electrode placement, mechanical fixturing, and geometry-aware calibration.

\subsection{Proof-of-concept on a complex shape}
\label{sec:complex_shape}

\begin{figure}[htbp] 
\centering
\includegraphics[width=0.48\textwidth]{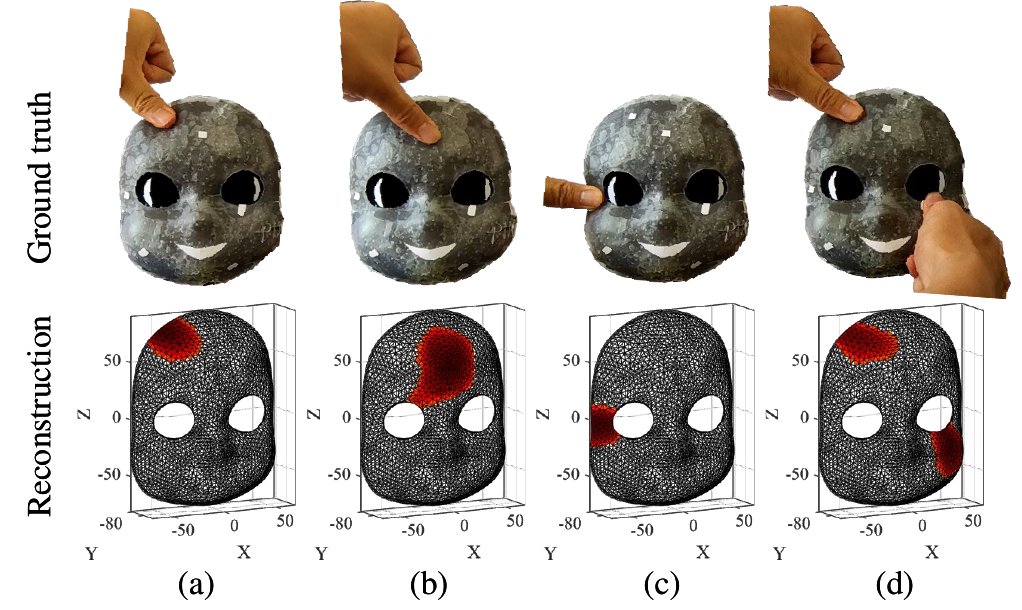} 
\caption{Example interactions with the iCub face sensor prototype. In each column, the ground truth (photography) is presented above the resulting EIT reconstruction. See the \href{https://drive.google.com/file/d/1e5cnhVCSGRQtdfwQZeAzJDwOmQ1GoHn_/view?usp=drive_link}{supplementary video} for more information.}
\label{fig:icub_face} 
\end{figure}

Finally, we explored the feasibility of applying the proposed conformal EIT sensing concept to a more complex freeform geometry. For this purpose, the front surface of the iCub robot's face was chosen. Unlike the planar and cylindrical prototypes used for quantitative evaluation, the iCub-face-shaped prototype required several fabrication modifications due to the high curvature and local geometric features of the surface. In preliminary trials, printing a 0.4~mm conductive TPU sensing layer on the face geometry has proven unreliable using a standard \ac{FDM} printer. The thin conductive layer was difficult to print continuously over the curved surface, and removing support material often led to damage of the sensing layer. Therefore, for this proof-of-concept prototype, the conductive TPU sensing layer was substituted with a 0.8~mm solid conductive TPU layer to improve printability and handling robustness. The compliant cover layer was also substituted. For this complex face geometry, fabricating a thin and conformal flexible cover layer using \ac{FDM} printing was impractical. Therefore, an soft cover layer was printed using \ac{SLA} to obtain a better surface fit and finer geometric resolution \cite{proper2023easy}. After manual assembly of boundary electrodes and conductive contact patches, manual touch tests were performed at representative locations on the face surface. As shown in Fig.~\ref{fig:icub_face}, the reconstructed conductivity-change regions generally appear near the touch locations. This result suggests that the proposed EIT sensing concept can be transferred from simple planar and cylindrical surfaces to more complex freeform geometry. 

This experiment serves as a qualitative demonstration of applying the proposed EIT sensing architecture to a humanoid-like freeform surface. A systematic quantitative evaluation on complex geometries would require further improvements in fabrication repeatability, electrode integration, contact-position control, and automated geometry-to-FEM generation. 

\section{Discussion}
The presented results demonstrate that tomographic tactile sensing and additive manufacturing are complementary approaches for humanoid skin design. EIT avoids dense internal taxel arrays by reconstructing contact-induced conductivity changes from boundary measurements, while 3D printing allows the sensing layer, support structure, and surface geometry to be designed for curved robot bodies. Strong material selection, electromechanical response, and geometry-aware reconstruction are important factors for large-area scaling to torso surfaces, hands, heads, and other non-developable geometries found on humanoid morphologies.

This work is a step toward deployable humanoid skin, but integration challenges regarding the conductive tape patches and externally assembled electrodes remain before the sensor can be truly fully 3D-printed. Future work should focus on fully printed conductive interfaces, integrated wiring, automated CAD-to-FEM generation, calibrated force-controlled testing on freeform surfaces, durability and hysteresis characterization, and integration with whole-body controllers for contact anticipation, impact mitigation, and safe physical interaction.

\section{Conclusion}
This work presented a 3D-printed conformal EIT tactile skin for contact localization on planar and curved surfaces. By combining a flexible conductive TPU sensing domain, conductive contact-enhancement patches, printed structural layers, and model-based EIT reconstruction, the sensor localizes contact without dense embedded taxels. Electromechanical characterization showed that contact-patch conductivity and the resistance/porosity of the conductive TPU sensing layer are key design parameters for improving sensitivity while preserving printability. Quantitative experiments demonstrated stable planar localization at moderate-to-high forces and 6~mm mean localization error on a curved U-shaped sensor, while a qualitative iCub-face prototype illustrated transfer to a more humanoid freeform geometry. These results indicate a practical path toward geometry-scalable tactile skins for humanoids operating in contact-rich human environments, including homes, care settings, and industrial workspaces.

\bibliographystyle{IEEEtran}
\bibliography{References2}
\end{document}